\UseRawInputEncoding
\documentclass[runningheads]{llncs}
\usepackage{times}
\usepackage{amsmath}
\usepackage{amsfonts}
\usepackage{graphicx}
\vspace{-0.1cm}  

\begin{document}
\title{Blockchain Phishing Scam Detection via Multi-channel Graph Classification }
%
%
\author{Dunjie Zhang\inst{1} \and
Jinyin Chen\inst{1,2}}
\authorrunning{D. Zhang et al.}
%
\institute{College of Information Engineering, Zhejiang University of Technology, Hangzhou 310023,China  \and
Institute of Cyberspace Security, Zhejiang University of Technology, Hangzhou,310023,China\\}
\maketitle              
\begin{abstract}
With the popularity of blockchain technology, the financial security issues of blockchain transaction network have become increasingly serious. Phishing scam detection methods will protect possible victims and build a healthier blockchain ecosystem. Usually, the existing works define phishing scam detection as a node classification task by learning the potential features of users through graph embedding methods such as random walk or graph neural network (GNN). However, these detection methods are suffered from high-complexity due to the large scale of the blockchain transaction network, ignoring temporal information of the transaction. Addressing to this problem, we defined the transaction pattern graphs for users and transformed the phishing scam detection into a graph classification task. To extract richer information from the input graph, we proposed a multi-channel graph classification model (MCGC) with a multiple feature extraction channels for GNN. The transaction pattern graphs and MCGC are more able to detect potential phishing scammers by extracting the transaction pattern features of the target users. Extensive experiments on seven benchmark and Ethereum datasets demonstrate that the proposed MCGC can not only achieve the state-of-the-art performance in graph classification task, but also achieve effective phishing scam detection based on the target users' transaction pattern graphs.

\end{abstract}
\section{INTRODUCTION}
\setlength{\parskip}{0\baselineskip}
As one of the most successful applications of blockchain\cite{2017The}, cryptocurrency\cite{yuan2018blockchain,wood2014ethereum} has promoted the rapid development of blockchain technology. The financial security\cite{chen2018detecting,bartoletti2020dissecting} of cryptocurrency has also become an important prerequisite for healthy development of blockchain technology. According to a report of \emph{Chainalysis}, 30,287 victims encountered financial scams on the Ethereum platform in the first half of 2017, including phishing scams, Ponzi schemes, ransomwares, etc., with a total economic loss of \$225 million\cite{EtherScamDB}. Among these scams, more than 50\% can be classified as phishing scams that take the cryptocurrency as the phishing target.

Traditional phishing scam detection methods\cite{khonji2013phishing} are usually applied to the identification of phishing emails or phishing webpages, to reduce the possibility of scams by warning users or directly blocking content. Due to the openness of the blockchain\cite{2017The}, the transaction records of all users are publicly available. By extracting the transaction records on blockchain, we can construct a large graph-structured transaction data, benefits from which it is possible to discover the identity features of different users from the blockchain transaction data through graph analysis methods\cite{wu2020comprehensive}. Modeling as different graph analysis tasks, such as node classification\cite{kipf2017semi,wang2020gcn}, graph classification\cite{10.5555/3327345.3327389,wang2016incremental}, and link prediction\cite{duan2017ensemble,fu2018link}, can help us identify potential scammers and provide assistance in solving financial scam on blockchain.

The existing works\cite{wu2019t,wu2020phishers,pareja2020evolvegcn,chenphishing} regard the phishing scam detection task as node classification. According to the transaction records between the target user and others, they extract the identity feature of the target user from the transaction information such as transaction address, transaction amount and timestamp. The unsupervised graph embedding methods\cite{wu2019t,wu2020phishers} based on random walk determine the process of random walk according to the transaction amount and timestamp. Based on graph neural network (GNN), \cite{pareja2020evolvegcn} and \cite{chenphishing} transform phishing scam detection task into a supervised dynamic node classification problem by learning the structural and dynamic features of the blockchain transaction network. It is worth noting that although these methods have achieved satisfying performance on blockchain phishing scam detection, they are suffered from high computational complexity since the node classification task often takes the transaction network containing all user nodes and transaction records as the input.

Considering the high-complexity of graph analysis on large-scale data, it is still a challenge to propose a low-complexity phishing scam detection model. From the perspective of the graph-structured data, the two nodes with a larger shortest path length can affect each other by transmitting messages through the edges. However, for the target node, it is often the node set in its neighborhood that has the greatest impact on it. Since the GNN-based graph classification methods\cite{10.5555/3327345.3327389,wang2016incremental} have achieved satisfying performance in the real-world datasets. It may be a feasible solution to construct transaction pattern graphs based on the neighbor transaction records of the target user. We transform the node classification task with the whole large-scale transaction network as input into a graph classification task with multiple small-scale transaction pattern graphs as input, which helps to achieve efficient phishing scam detection. The main contributions of our work are summarized as follows:

\setlength{\parskip}{0\baselineskip}
\setlength{\hangindent}{2.3em}
$\vcenter{\hbox{\tiny$\bullet$}}$
We firstly defined the transaction pattern graphs for blockchain transaction users. Each user has its own small-scale transaction pattern graph, which makes it possible to detect potential phishing scammers with less computational complexity through graph classification.

\setlength{\hangindent}{2.3em}
$\vcenter{\hbox{\tiny$\bullet$}}$
We proposed a multi-channel graph classification model, namely MCGC. The proposed MCGC has a multi-channel GNN architecture, which can automatically extract richer information from the different pooling graphs.

\setlength{\hangindent}{2.3em}
$\vcenter{\hbox{\tiny$\bullet$}}$
Extensive experiments conducted on seven benchmark and an Ethereum datasets demonstrate that the MCGC achieves the state-of-the-art performance in graph classification task and can effectively detect the phishing nodes according to the transaction pattern graphs.

\section{RELATED WORK}
Our work builds upon two categories of recent research: phishing scam detection and graph classification.

\subsection{Phishing Scam Detection}
Phishing scam detection methods can identify phishing scammers before the scam occurs, or provide early warning for possible victims. In this section, we briefly review phishing scam detection methods, mainly categorized into random walk-based methods and GNN-based methods.

\textbf{\emph{Random walk-based methods.}} Wu et al.\cite{wu2020phishers} proposed a novel network embedding algorithm called trans2vec, which is composed of random walk sampling and node sequence embedding. According to the transaction amount and timestamp, trans2vec performs biased random walk in the network to obtain a large number of node sequences, which are used to extract the users' node features. Wu et al.\cite{wu2019t} further proposed T-EDGE, whose main structure is similar to the main structure of trans2vec. The main improvement of T-EDGE is to ensure the sequence of nodes during the random walk, and it also solves the problem of multiple edges in the transaction network.

\textbf{\emph{GNN-based methods.}} To make full use of the powerful learning representation ability of GNN, Pareja et al.\cite{pareja2020evolvegcn} proposed EvolveGCN, which adapts the graph convolutional network (GCN) model along the temporal dimension without resorting to node embeddings. EvolveGCN captures the dynamism of the graph sequence through using an recurrent neural network (RNN) to evolve the GCN parameters. Tam et al.\cite{handason2019identifying} proposed a new message passing mechanism named EdgeProp, which allows multi-dimensional continuous edge features propagating into node embeddings when performing node classification task.

In summary, whether it is a random walk or a GNN-based method, existing work regards phishing scam detection as a node classification task. When applied to the blockchain transaction network, the huge number of nodes and transaction records lead to the high-complexity of these detection methods.

\subsection{Graph Classification}
Graph classification can predict the graph-level labels of small-scale graphs, which has achieved satisfying performance in real-world datasets such as bioinformatics and chem\\-otherapy informatics. Here we briefly introduce graph classifiers.

Narayanan et al.\cite{narayanan2017graph2vec} proposed a neural embedding framework named Graph2vec to learn data-driven distributed representations of arbitrary sized graphs. Graph2vec can solve the problem of poor generalization ability of graph kernel methods\cite{yanardag2015deep,vishwanathan2010graph} in an unsupervised manner. Ying et al.\cite{10.5555/3327345.3327389} proposed a differentiable graph pooling method named Diffpool, which aggregates the nodes into a new cluster as the input of the next layer by the cluster assignment matrix. Due to the high complexity of the learning cluster assignment matrix, the self-attention graph pooling (SAGPool)\cite{lee2019self} considers both node features and graph topology. SAGPool selects top-K nodes based on a self-attention mechanism to form the induced subgraph for the next input layer. OTCOARSENING proposed by Ma et al.\cite{ma2019unsupervised} designs a coarsening strategy based on hierarchical abstraction through minimizing discrepancy along the hierarchy, which can be combined with unsupervised learning methods. GMN\cite{khasahmadi2020memory} introduces an efficient memory layer for GNNs that can jointly learn node representations and coarsen the graph.

\section{PRELIMINARY}
This section introduces the problem definition of the graph classification and the phishing scam detection on blockchain.
We represent a graph as $G=(V,E,A)$, where $V$ is the node set with $|V|=N$, $e_{i,j}=<v_i,v_j> \in E$ is the edge between the node $v_i$ and $v_j$.  $A\in \mathbb{R}^{N \times N}$ is the adjacency matrix, where $A_{i,j}\neq0$ denotes node $v_i$  directly connected with $v_j$ while $A_{i,j}=0$ otherwise. The graph $G$ may contain an attribute vector of each node in some case, here we denote the attribute vector of graph $G$ as $A\in \mathbb{R}^{N \times D}$, where $D$ is the dimension of $X$. Generally, the adjacency matrix $A$ contains the information of $V$ and $E$ on graph $G$, so we use $G=(A,X)$ to represent a graph more concisely.

\setlength{\parskip}{0.5\baselineskip}
\noindent\textbf{DEFINITION 1 (Graph classification).} For a graph classification dataset $G_{set}$, it includes $M$ graphs $\{G_1, G_2,..., G_M\}$.  The graph classification task aims to predict the categories of unlabeled graph $G_{u} \subset G_{set}$ through the model $f_{\theta}^{graph}(\cdot)$  trained by the labeled graphs $G_{l} = G_{set}-G_{u}$ with its corresponding label $Y=\left[y_{1}, \cdots, y_{|G_{l}|}\right]$.

\noindent\textbf{DEFINITION 2 (Phishing scam detection on blockchain).} On blockchain transaction network, the node set $V$ represents the users of the blockchain trading platform, and $E$ represents the transaction record set between different users. Here, there may be $n$ transactions $\{e_{i,j}^0, e_{i,j}^1,...,e_{i,j}^n\}$ between users $v_i$ and $v_j$. The phishing scam detection task aims to predict the categories of unlabeled nodes $V_{u} \subset V$ $f_{\theta}^{phishing}(\cdot)$ trained by the labeled nodes $V_{l} = V-V_{u}$ with its corresponding label $F=\left[\tau_{1}, \cdots, \tau_{l}\right]$, where $\tau_{i}$ is the ground truth label of $v_i$. $\tau_{i}=1$ denotes the node $v_i$ is a phishing node while $\tau_{i}=0$ otherwise.
\setlength{\parskip}{0\baselineskip}

\section{METHODOLOGY}
To transform the high-complexity node classification task into a lower-complexity graph classification task on blockchain transaction network, we define the user transaction pattern graphs for the first time. Specifically, we take the whole graph-level representation of the transaction pattern graph as the target user's transaction pattern feature, which provides a more efficient method for phishing node detection. Additionally, to extract richer information from the transaction pattern graphs, we propose MCGC, a graph classification model with a multi-channel GNN architecture, which aggregates the information of pooling graphs in multiple channels in a trainable manner, thus achieving better graph classification performance.

\subsection{Transaction Pattern Graph Construction}\label{4.1}
\begin{figure}[htbp]\setlength{\belowcaptionskip}{-0.5cm}
  \centering
  \includegraphics[width=1\linewidth]{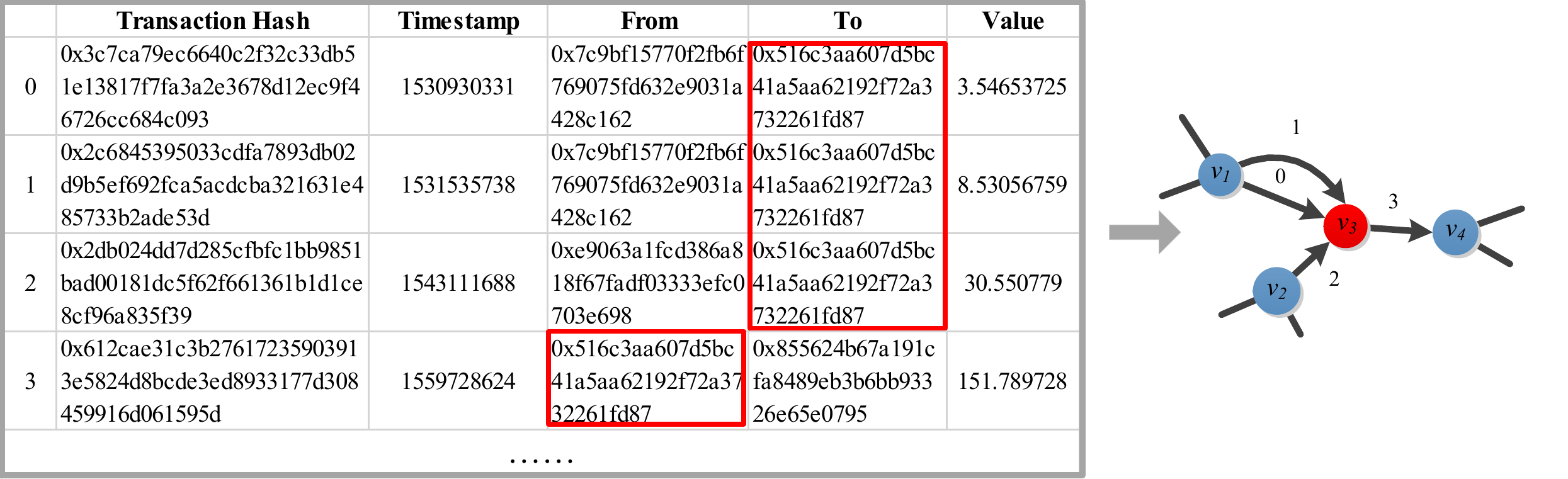}\\
  \caption{The process of Ethereum transaction pattern graph construction.}
  \label{fig1}
\end{figure}

In this part, we introduced the construction process of the transaction pattern graphs for the Ethereum trading platform. We collected Ethereum transaction data from the Ethereum trading platform (https://etherscan.io/) through Ethereum clients Geth and Parity. Each transaction data in this website contains dozens of attributes, among which the transaction timestamp, transaction sending and receiving address, and transaction amount are the key information for constructing a transaction network. The sending address and receiving address correspond to the nodes in the transaction network, the transaction timestamp and the transaction amount indicate the existence edges and their information between the corresponding node pairs. We reserve these transaction information from the original data to construct the transaction pattern graphs. The partial transaction records and the process of constructing a transaction pattern graph are shown in Fig.\ref{fig1}. Fig.\ref{fig1}(a) shows the partial transaction records of the target node $v_0$ that marked in red, and Fig.\ref{fig1}(b) shows the construction process of the first-order transaction pattern graph of $v_0$. The 4 transaction records in Fig.\ref{fig1}(a) contain 4 different addresses, corresponding to 4 edges and 4 nodes in the transaction graph.

Through the above steps, we took the target node $v_0$ as the central node, and extracted the user address of the other party as the first-order transaction node based on the transaction records. Then, the first-order transaction nodes are regarded as the central nodes, the second-order transaction nodes are extracted according to the same method. Repeat this step until the designated $K$-order transaction pattern graph is constructed. Different from a large-scale Ethereum transaction network used for node classification task, we constructed independent small-scale transaction pattern graph for each node. There may be multiple transaction records between two transaction addresses on blockchain transaction data, In this case, we merge multiple transaction records into one transaction, taking the summed transaction amount as a new transaction amount information, and the average timestamp as the new edge information.

Here, we choose the 1259 phishing nodes marked in \cite{wu2020phishers} and the same number of active normal nodes randomly selected in the same period as our target node set. Besides, we set $K=4$ to ensure that the transaction pattern graphs still contain enough information after merging multiple transaction records.

\subsection{Multi-channel Graph Classification}

\begin{figure}[htbp]\setlength{\belowcaptionskip}{-1cm}
  \centering
  \includegraphics[width=1\linewidth]{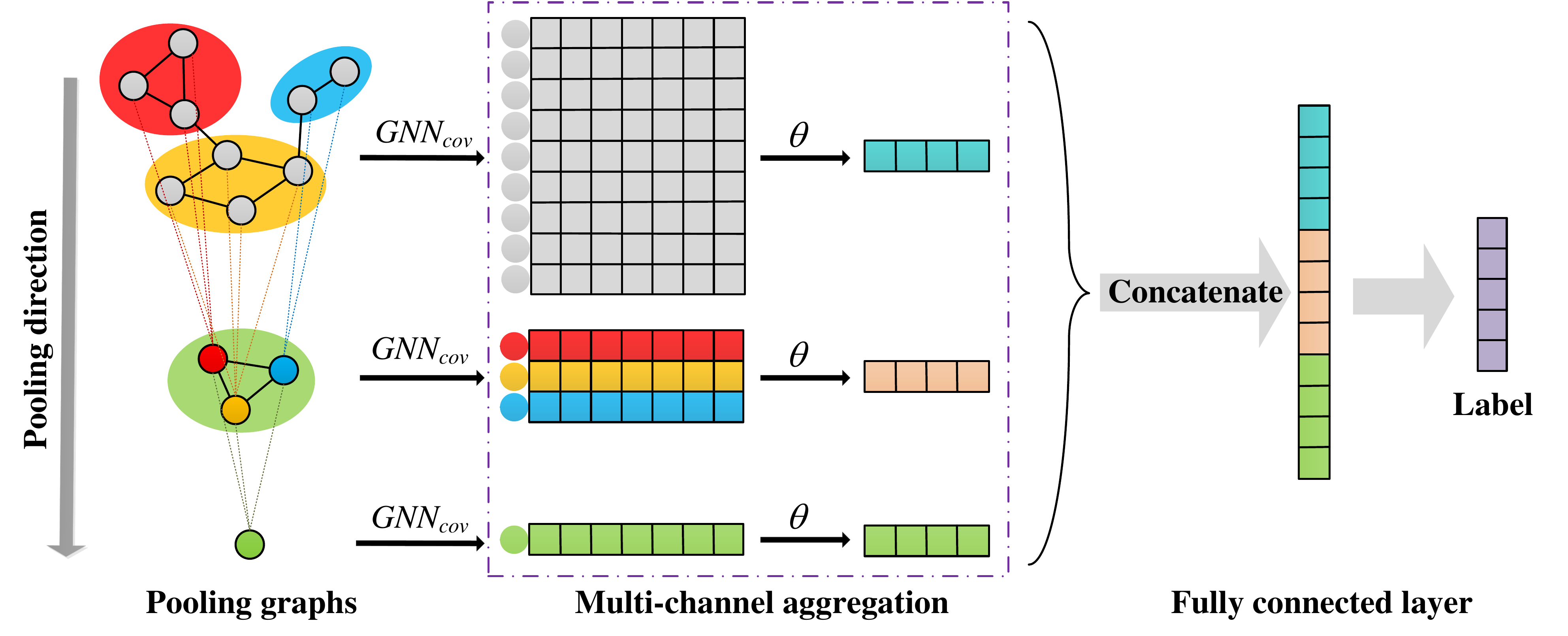}\\
  \caption{The specific illustration of multi-channel architecture in MCGC.}
  \label{figflow}
\end{figure}

Our proposed approach, MCGC, utilizes a multi-channel architecture to fuse the node-level representations of different pooling graphs. The specific architecture of multi-channel architecture is shown in Fig.\ref{figflow}. The key intuition is that for each hierarchical graph pooling layer of MCGC, we introduce a trainable node importance weights to aggregate the node-level representations to the graph-level. Thus, we can capture the important structural information for graph classification from different pooling graphs. After that, MCGC learns effective features from these structural information and outputs the final graph classification results.

\subsubsection{Hierarchical graph pooling.} MCGC aggregates the nodes in the input graph into a new node cluster through the node aggregation operation of the graph pooling layer, thus obtaining the coarsened pooling graph structure. Multiple graph pooling layers can extract different crucial hierarchical structure information of the input graph, which helps us extract multi-level user transaction pattern information. In addition, the pooling layer can be combined with GNNs to form an end-to-end training model, and has demonstrated good performance on many real-world graph classification datasets.

Specifically, MCGC employs the propagation function to implement the convolution layer for extracting the $l$-th node-level representation, which can be expressed as:
\begin{equation}
\label{eq1}
  H^{l,k}=\sigma({\hat{A}^l} H^{l,k-1} W^{l,k})
\end{equation}
where $\hat{A}^l=\tilde{D}^{-\frac{1}{2}} \tilde{A}^l \tilde{D}^{-\frac{1}{2}}$, $A^l$ is the $l$-th adjacency matrix of input graph $G$, and $\tilde{A^l}=A^l+I^l_{N}$ is the adjacency matrix of the $l$-th pooling graph $G^l$ with self-connections.  $I^l_{N}$ is the identity matrix and $\tilde{D}_{i i}=\sum_{j} \tilde{A^l}_{i j}$  denotes the degree matrices of $\tilde{A^l}$. $W^{l,k}$ is the parameters of the $k$-th propagation function in the $l$-th architecture. $\sigma$ is the Relu active function. When $k=1$, $H^{l,0}$ is the node attribute $X^l$ of the $l$-th architecture.

The $(l+1)$-th node-level representation $H^{l+1}$ is usually obtained  by running $K$ iterations of Eq.\ref{eq1}.  To achieve effective graph convolution with lower complexity, here we choose $K=3$. The convolution layer of the $l$-th architecture is denoted as:

\begin{equation}
\label{eq2}
  H^{l+1}=GNN_{cov}(A^{l},X^{l};\theta^{l})
\end{equation}
where $H^{l+1}$ is computed computed from the $l$-th adjacency matrix $A^l$ and node attribute $X^{l}$. $\theta^l$ is the parameters set of the $l$-th architecture. The input $A^{0}$ and $X^{0}$ are the original adjacency matrix and node attributes on graph, i.e., $A^{0}=A$, $X^{0}=X$, respectively.

 For the pooling layer, it calculates the cluster assignment matrix $C^{l} \in \mathbb{R}^{n_{l} \times n_{l+1}}$ according to the topological structure and node representation of the current graph.  The cluster assignment matrix of the $l$-th architecture is:
\begin{equation}
\label{eq3}
  C^l=softmax \left ( GNN_{pool}(A^l,X^l) \right )
\end{equation}
where $GNN_{pool}$ has the same structure as $GNN_{cov}$. $n_l$ and $n_{l+1}$ represent the number of nodes in the $l$-th and the $(l+1)$-th pooling graphs, respectively . Each row in the $C^{l}$ represents the probability that the node is assigned to each node cluster in the $l+1$-th layer. Generally, $n_l>n_{l+1}$, the pooling layer can coarsen the graph into a pooling graph with a smaller number of nodes, which is helpful for extracting the crucial structure of the original graph.

According to the adjacency matrix $A^l$ and node attribute $X^l$ of the $l$-th pooling graph, the convolutional layer and the pooling layer obtain the node-level representation $H^l$ and cluster assignment matrix $C^l$ of the $l$-th layer, respectively. Then the pooling adjacency matrix $A^{l+1}$ and the node attribute $X^{l+1}$ of $(l+1)$-th layer are calculated by:
\setlength{\parskip}{0\baselineskip}
\begin{equation}
\label{eq4}
  X^{l+1}={C^{l}}^T H^l \in \mathbb{R}^{n_{l+1} \times d}
\end{equation}

\setlength{\parskip}{-0.8\baselineskip}
\begin{equation}
\label{eq5}
  A^{l+1}={C^{l}}^T A^l C^{l} \in \mathbb{R}^{n_{l+1} \times n_{l+1}}
\end{equation}
where $d$ denotes the feature dimension of each node. Eq.~\ref{eq4} aggregates the $H^l$ according to the cluster assignments $C^l$, generating a new pooling node attribute for each of $n_{l+1}$ clusters. Similarly, Eq.~\ref{eq5} generates a new pooling adjacency matrix based on the adjacency matrix $A^l$, denoting the connectivity strength between each pair of clusters.

\setlength{\parskip}{-0.5\baselineskip}
\subsubsection{Multi-channel structure.} MCGC extracts $L$ pooling graphs $\{G^1(A^1,X^1) ,..., G^L(A^L,\\X^L)\}$ from the original graph $G(A, X)$ through $L$ hierarchical graph pooling layers. The topological structure and node-level representations of these pooling graphs reflect the multiple channel representations of $G$.  Intuitively, the crucial graph structures learned by the pooling graphs of different channels are also different. We hope to capture the relationship between different channels and graph-level representations in a learnable manner, instead of simply using mean-pooling or max-pooling to obtain graph-level representation.
\setlength{\parskip}{0\baselineskip}
For the above considerations, we introduce a trainable node importance weights for each pooling graph to learn its graph-level representation. Specifically, for the $l$-th pooling graph, MCGC obtains its node-level representation through the graph convolutional layer $GNN_{cov}$. The importance values of different nodes in the current pooling graph are learned by the proposed trainable node importance weights, and then we aggregate the node-level representation of the pooling graph into the graph-level by weighted summation, which can be expressed as:
\begin{equation}
\label{eq6}
  S_l = \frac{\sum {\theta_l(i) \cdot Z_l(i)}}{\sum {\theta_l}}
\end{equation}
where $S_l \in \mathbb{R}^d$ denotes the graph-level representation of the $l$-th pooling graph.

Then, we combine the graph-level representations of the original graph and the pooling graphs. MCGC preserves the multi-channel information of the graph as much as possible by concatenating the graph-level representations of different pooling graphs together. The global-level representation $S \in \mathbb{R}^{(L+1)d}$ is denotes as:
\begin{equation}
\label{eq7}
  S=concat(normal(S_0),...normal(S_L))
\end{equation}
where $concat(\cdot)$ denotes the concatenate function, which stitches the graph-level representations of the original graph and the $L-1$ pooling graphs. $normal(S_l)=S_l/\sum S_l$ is a normalization function, which converts the graph-level representation of each pooling graph into a identity vector. It can avoid the inconsistency of the scope of graph-level representations caused by the difference of the number of nodes.

Finally, we get the prediction probability of the input graph through the fully connected layer with a softmax classifier:
\begin{equation}
\label{eq8}
  O=softmax(SW+b)
\end{equation}
where $ W\in \mathbb{R}^{(L+1)|Y|}$ and $b \in \mathbb{R}^{|Y|}$ denote the weight and bias of the fully connected layer, respectively. $|Y|$ is the number of labels in the graph dataset $G_{set}$.

\subsubsection{Training procedure.} The entire MCGC is an end-to-end model that can be trained by stochastic gradient descent. For a set of graphs $G_{set}$, we employ the following loss function $\mathcal{L}$ to train our MCGC, which can be represented as:
\begin{equation}
\label{eq9}
  \mathcal{L} = -\sum_{G_i \in G_{set}}^{|G_{set}|} \sum_{j=1}^{|Y|}Q_{ij}\ln{O_{ij}\left(A_i,X_i\right)}+\sum_{l=1}^L \sum_{i=1}^{n_l} \frac{1}{n_l}H(C_l(i))
\end{equation}
where $Y=\left\{ y_{1},...,y_{|Y|}\right\}$ is the category set of the graphs, $Q_{ij}$ is the ground truth with $Q_{ij}=1$ if graph $G_i$ belongs to category $y_{j}$ and $Q_{ij}=0$ otherwise. $O_{ij}$ denotes the predicted probability that graph $G_i$ belongs to $y_{j}$, which is calculated by Eq.\ref{eq8} and can be considered as a function of $A_i$ and $X_i$, thus we denote it as $O_{ij}\left(A_i,X_i\right)$. $H(\cdot)$ denotes the information entropy function.

$\mathcal{L}$ consists of two parts. The first part represents the cross-entropy of the classification prediction probability and the ground truth label, which can guide the prediction probability to be closer to the ground truth label.  The second part represents the information entropy constraint of the cluster assignment matrix of the $l$-th layer, which helps the row vector of the cluster assignment matrix to approach the ont-hot vector and better learn the mapping relationship between nodes.

\section{EXPERIMENTS}
To verify the performance of MCGC, we conduct graph classification experiments on several benchmark graph classification and an Ethereum transaction datasets. For each dataset, we perform 10-fold cross-validation and report the average accuracy. In our MCGC, we implement three hierarchical graph pooling layers, i.e., we set $L=3$. We use the Adam optimizer to optimize the model, and the learning rate is searched
in {0.1,0.01,0.001}. The feature dimension $d$ is set by a hyper-parameter search in $\{32,64,12\\8,256,512\}$. We implement our proposed MCGC with PyTorch, and our experimental environment consists of i7-7700K 3.5GHzx8 (CPU), TITAN Xp 12GiB (GPU), 16GBx4 memory (DDR4) and Ubuntu 16.04 (OS).

\subsection{Datasets}
To verify whether the multi-channel structure can better aggregate node-level representation and the phishing node detection performance of MCGC, we select seven benchmark and an Ethereum transaction datasets for our graph classification experiments. Among the benchmark datasets, two datasets are social network datasets, including IMDB-BINARY and REDDIT-BINARY. The others are about bio-informatics and chemo-informatics. Each dataset is composed of two classes of graphs. The basic statistics are summarized in Table \ref{tab:data}.

\subsection{Baselines}
To verify the performance of MCGC, we compare it with 5 advanced graph classification methods including Graph2vec\cite{narayanan2017graph2vec}, Diffpool\cite{10.5555/3327345.3327389}, SAGPool\cite{lee2019self}, OTCOARSENING\cite{ma2019unsupervised} and GMN\cite{khasahmadi2020memory}. The specific method is introduced as follows:

\noindent\textbf{Graph2vec.} It establishes the relationship between a network and the rooted subgraphs. It extracts rooted subgraphs and provides corresponding labels into the vocabulary, and then trains a skipgram model to obtain the representation of the network.

\noindent\textbf{DIFFPOOL.} It learns a differentiable soft cluster assignment for nodes at each layer, and generates hierarchical representations of graphs in an end-to-end manner.

\noindent\textbf{SAGPool.} Based on self-attention, It uses graph convolution to capture both node features and graph topology. Compared with DIFFPOOL, this algorithm can learn hierarchical representations of graphs with fewer parameters.

\noindent\textbf{OTCOARSENING.} It designs a coarsening strategy based on hierarchical abstraction through minimizing discrepancy along the hierarchy, which can be combined with unsupervised learning methods.

\noindent\textbf{GMN.} It designs an efficient memory layer for GNNs that can jointly learn node representations and coarsen the graph. It consists of a multi-head array of memory keys and a convolution operator to aggregate the soft cluster assignments from different heads.


\begin{table}[]
\centering
\caption{The basic statistics of eight datasets.}
\scalebox{1.0}{
\begin{tabular}{c|cccc}

\hline \hline
Dataset       & \#Graphs & \#Classes & \#Ave\_nodes    & \#Ave\_edges    \\ \hline
MUTAG   \cite{debnath1991structure}      & 188      & 2         & 17.92              & 20.42         \\
PTC \cite{toivonen2003statistical}          & 344      & 2         & 14.29             & 14.69         \\
PROTEINS \cite{borgwardt2005protein}     & 1113     & 2         & 39.06             & 72.82         \\
NCI1   \cite{shervashidze2011weisfeiler}       & 4110     & 2         & 29.87             & 32.30         \\
NCI109   \cite{shervashidze2011weisfeiler}     & 4127     & 2         & 29.69             & 32.13         \\
IMDB-BINARY \cite{yanardag2015deep}  & 1000     & 2         & 19.77             & 96.53         \\
REDDIT-BINARY \cite{yanardag2015deep}& 2000     & 2         & 429.63            & 497.75        \\
Ethereum      & 2518     & 2         & 120.43            & 130.08        \\ \hline
\hline
\end{tabular}
}
\label{tab:data}
\end{table}

\subsection{Evaluation Metrics} The datasets in our experiment are all binary datasets, in which the number ratio of positive examples to negative examples tends to 1. Therefore, we only use the accuracy to evaluate the performance of different algorithm, which can be expressed as:
\begin{equation}
\label{eq10}
  Accuracy=\frac{TP+TN}{TP+TN+FP+FN}
\end{equation}
where TP and TN are the number of positive and negative examples predicted correctly, respectively. FP and FN are the number of positive and negative examples predicted incorrectly, respectively.

\subsection{Graph Classification Performance }
To better detect phishing nodes on blockchain transaction network through the powerful learning representation ability of GNN, we transform the phishing detection task into a graph classification task with a smaller graph scale and lower training complexity. We first verify the performance of the proposed MCGC in the benchmark graph classification datasets. Compared with the baselines in Table \ref{tab:performance}, the proposed MCGC achieves state-of-the-art performance among all benchmark datasets. Additionally, the performance improvement of MCGC on the REDDIT-BINARY is more obvious. This may be due to the most intuitive connection between the label of REDDIT-BINARY and the topological structure of the graph. The graph structure generated by Q \& A interaction is usually similar to a star network, while the graphs of user discussion interaction usually have no obvious central node.

\begin{table}[]
\centering
\caption{The graph classification performance of seven datasets by various methods.}
\scalebox{0.95}{
\begin{tabular}{c|ccccccc}
\hline

\hline
Methods       & MUTAG    & PTC      & PROTEINS   & NCI1    &NCI109    &IMDB-BINARY      &REDDIT-BINARY    \\ \hline
Graph2vec     & 83.15    & 61.59    & 73.30      & 73.22   &  74.26   &62.74            &59.07            \\
Diffpool      & 80.60    & 62.00    & 75.90      & 74.29   &  74.10   &75.20            &86.19            \\
SAGPool       & 78.60    & 61.39    & 73.30	     & 74.18   &74.06	  &72.20            &73.90        \\
OTCOARSENING  &85.60     & 63.57	& 74.90	     & 76.18   & 68.50	  &74.60	        &76.53     \\
GMN        &90.53	&64.59	&75.78	&73.17	&73.26	&77.00	&87.36    \\
MCGC   &91.67	&64.71	&78.91	&76.54	&76.36	&77.00	&91.00       \\            \hline

\hline
\end{tabular}
}
\label{tab:performance}
\end{table}

Generally, the graph classification performance of various graph classification methods is different on bioinformatics datasets and social network datasets. For bioinformatics  datasets, the classification results obtained by different methods are relatively close. Since MUTAG, PTC, PROTEINS, NCI1, and NCI109 are all small-scale graphs, they usually only contain dozens of nodes. The simple topology makes it easy to extract the structure information of these graphs. However, the hierarchical representation learning of graph begins to show its advantages for social network datasets with complex structure. The classification accuracy of Diffpool and other hierarchical graph classification methods on IMDB-BINARY and REDDIT-BINARY is 8\%-10\% higher than Graph2vec. Since MCGC is based on the hierarchical structure of Diffpool, it aggregates multi-channel hierarchical graph structure information in a learning manner, which helps to better aggregate the node-level representation of hierarchical pooling graphs into the graph-level, thus obtaining better graph classification performance.

\subsection{Phishing Node Detection on Blockchain Network }
\begin{figure}[htbp]\setlength{\belowcaptionskip}{-0.5cm}
  \centering
  \includegraphics[width=0.7\linewidth]{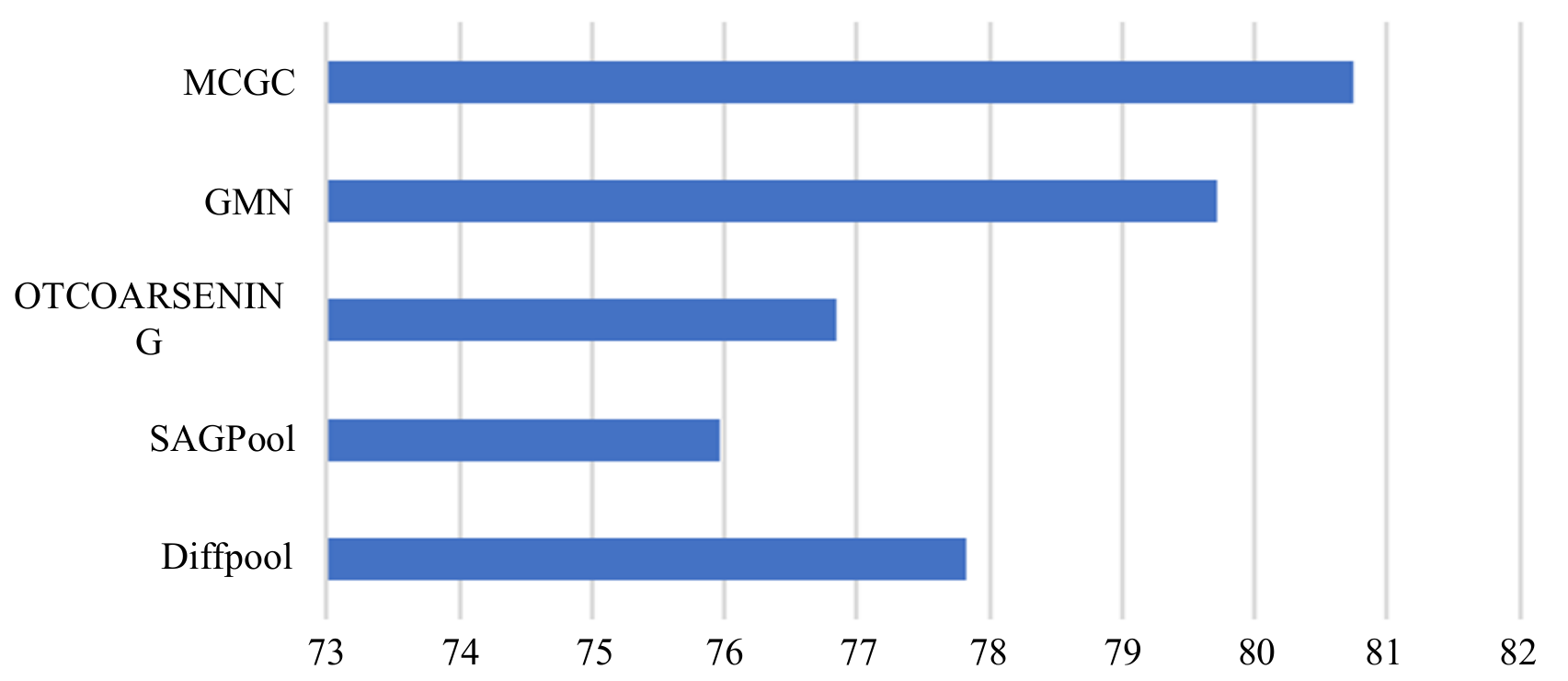}\\
  \caption{The graph classification accuracy of Ethereum dataset.}
  \label{fig.2}
\end{figure}

The blockchain account information based on the hash value of the public key makes users further obscure the identity attribute information on the basis of pseudonyms. This makes it difficult to define a user's identity features based on information other than his account transaction records. Considering that it is still a huge challenge to analyze large-scale data in deep GNN models, we transform the phishing scam detection into a small-scale graph classification task. We learn the potential features of the users by building transaction pattern graphs centered on these user nodes. In this part, we conduct the graph classification experiments on the transaction pattern graphs of phishing and normal nodes. The performance of the proposed MCGC and baselines are shown in Fig.\ref{fig.2}. We can see that in the Ethereum transaction network, MCGC can still achieve state-of-the-art performance, which indicates that the proposed MCGC can better extract the potential identity features of the transaction users.

\begin{figure}[htbp]\setlength{\belowcaptionskip}{-1cm}
  \centering
  \includegraphics[width=1\linewidth]{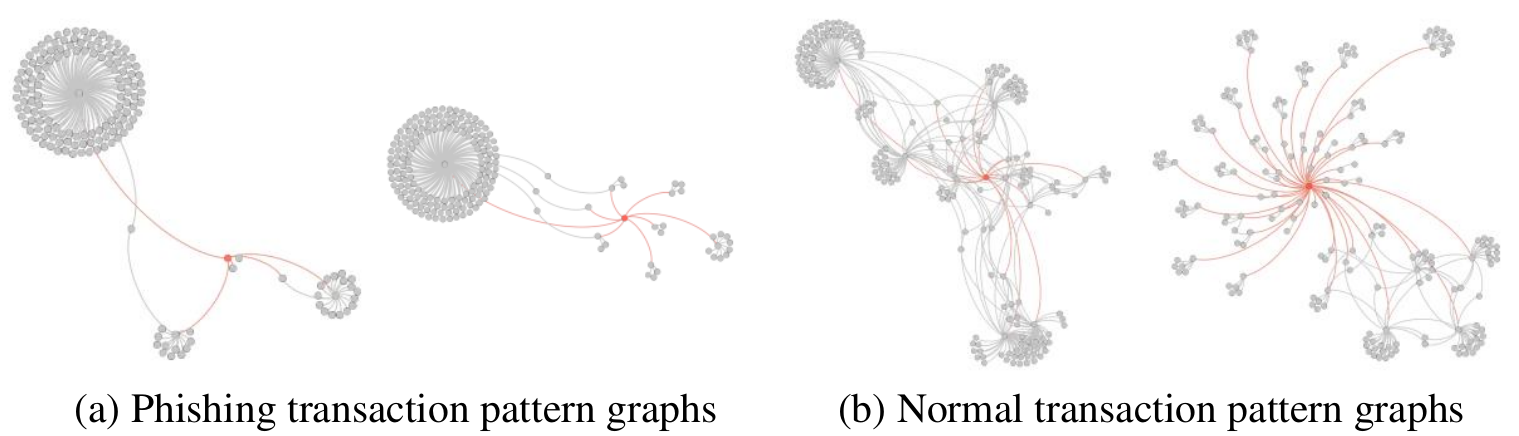}\\
  \caption{The visualization of phishing and normal transaction mode graphs.}
  \label{fig.3}
\end{figure}

To further investigate the reason why the hierarchical graph classification methods perform well on the Ethereum dataset, we visualize the topological structure of the transaction pattern graphs of phishing nodes and normal nodes, respectively. Fig.\ref{fig.3} (a) are two transaction pattern examples of phishing nodes, and (b) are two transaction pattern examples of normal nodes. The red node is the target user, the gray node is its neighboring node, the red edge are the transaction records of the target user, and the gray edges are the transaction records between neighboring nodes. Compared with normal users, phishing users have fewer direct trading users. The nodes directly connected to the phishing nodes often have a larger node degree value. They are usually exchange addresses, which are used for asset management, currency exchange, and so on.

Hierarchical graph classification methods such as MCGC can effectively aggregate the nodes with a larger node degree value and their neighbor node sets. Normal nodes are more likely to be aggregated, while phishing nodes tend to exist as independent nodes in the next pooling graph. When the node-level representations are aggregated into the graph-level, the feature vector of the phishing node tends to account for a larger proportion, which instructs the graph classifier to classify the input graph into the phishing transaction pattern class.

\subsection{Time Efficiency of Phishing Detection}
In this part, we verify the time efficiency of the GNN-based phishing detector and MCGC on the blockchain transaction network. Since EvolveGCN\cite{pareja2020evolvegcn} and EdgeProp\cite{handason2019identifying} consider dynamic information or edge information based on GCN, respectively. They have higher complexity than the traditional GCN. Therefore, we take the simplest GCN\cite{kipf2017semi} as an example to detect phishing scam on the blockchain transaction network. Specifically, we select multiple nodes as the central nodes at the same time, and construct the whole $K$-order transaction graph by the construction process in Section~\ref{4.1}, which is used in the GCN-based phishing scam detector.

\begin{figure}[htbp]\setlength{\belowcaptionskip}{-0.5cm}
  \centering
  \includegraphics[width=0.7\linewidth]{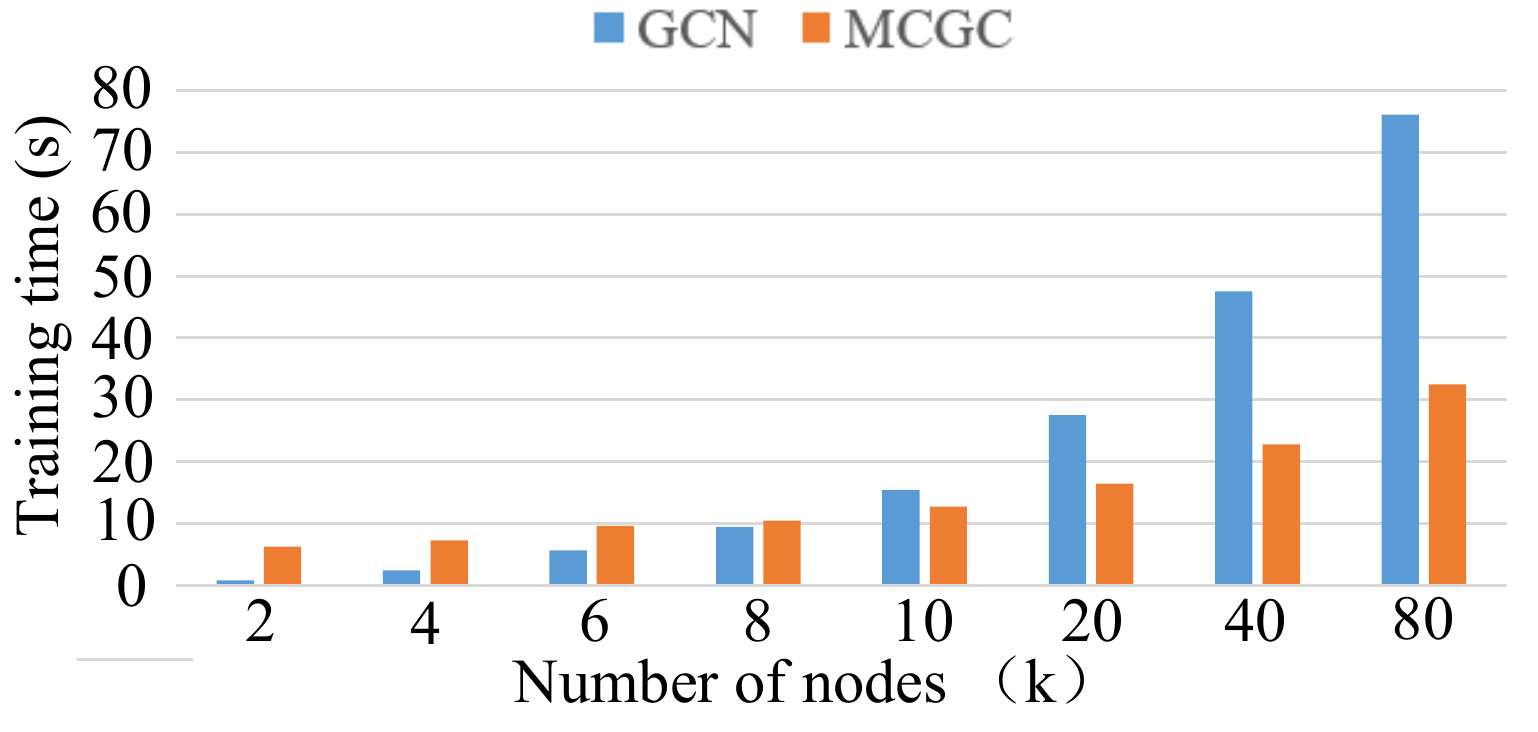}\\
  \caption{The training time in each iteration when detecting different sizes of transction network.}
  \label{fig5}
\end{figure}

Fig.\ref{fig5} shows the time efficiency of phishing scam detection under different node scales for GCN and MCGC. When the number of nodes is small, GCN has higher time efficiency. However, as the number of nodes increases, the computational complexity of GCN increases rapidly. When the number of nodes exceeds 40k, the complexity of MCGC is relatively lower. In addition, facing users newly added to the transaction network, the phishing scam detectors based on node classification may need to reconstruct the transaction graph for retraining, while our MCGC can detect any newly added user transaction pattern graph without retraining.

\section{CONCLUSION}
In this paper, we firstly define the transaction pattern graphs for blockchain transaction data, and propose a graph classification model with a multi-channel GNN architecture, named MCGC. To reduce the computational complexity of the phishing scam detection, we transform the blockchain phishing scam detectopm into a graph classification task, and build the independent transaction pattern graphs for the target blockchain transaction users. Moreover, to extract richer transaction pattern features from the transaction pattern graphs, we regard the pooling graphs learned from different hierarchical graph pooling layers as the multi-channel representations, and introduce a trainable node importance weights to better aggregate the information of multi-channel pooling graphs. Experiments on seven benchmark and an Ethereum datasets demonstrate that MCGC can not only achieve the state-of-the-art performance in graph classification task, but also achieve effective phishing scam detection based on the target users’ transaction pattern graphs.

%
%
%
\bibliographystyle{IEEEtran}
\bibliography{refer}

%
%
%
%
\end{document}